\documentclass[10pt,twocolumn,letterpaper]{article}

\usepackage[pagenumbers]{cvpr} 

\definecolor{cvprblue}{rgb}{0.21,0.49,0.74}
\usepackage[pagebackref,breaklinks,colorlinks,allcolors=cvprblue]{hyperref}

\def\paperID{7284}
\def\confName{CVPR}
\def\confYear{2025}

\title{SfM-Free 3D Gaussian Splatting via Hierarchical Training}

\author{Bo Ji \qquad Angela Yao \\
National University of Singapore\\
{\tt\small \{jibo,ayao\}@comp.nus.edu.sg}
}

\usepackage{graphicx}
\usepackage{booktabs}
\usepackage{makecell}
\usepackage{multicol}
\usepackage{multirow}

\usepackage{eurosym}
\usepackage{comment}
\usepackage{bbding,pifont}
\usepackage{soul}
\usepackage{tabularx}
\usepackage{pifont}
\usepackage{subcaption}
\usepackage{wrapfig}

\usepackage{amsmath}
\usepackage{amssymb}
\usepackage{booktabs}
\usepackage{colortbl}

\usepackage{svg}
\usepackage{algpseudocode}

\definecolor{bgcolor}{rgb}{0.9,0.9,0.9}

\begin{document}

\twocolumn[{%
    \renewcommand\twocolumn[1][]{#1}%
    \maketitle
    \centering
    \includegraphics[width=0.95\linewidth]{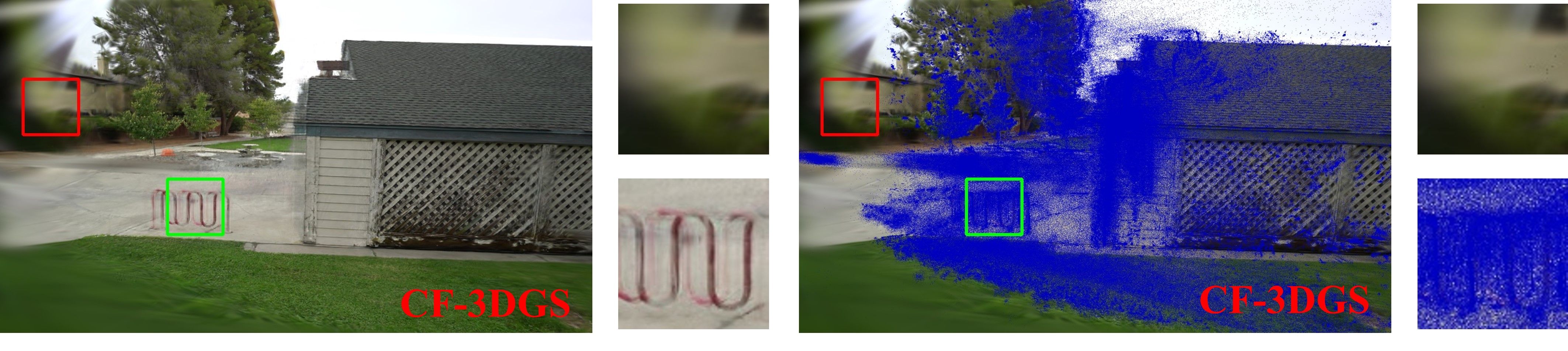}
    \includegraphics[width=0.95\linewidth]{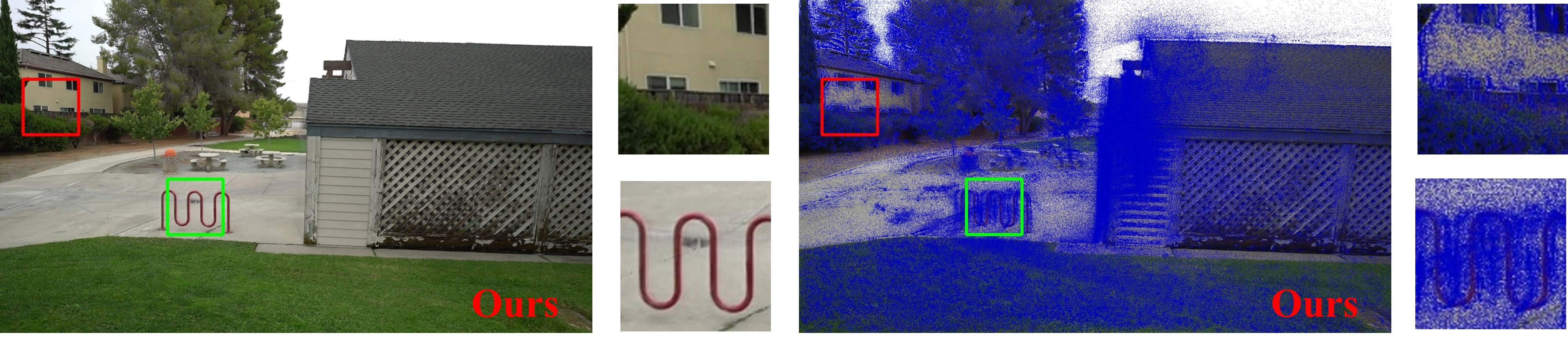}
	
    \captionof{figure}{
    \textbf{Novel view synthesis results (left) alongside the projected centers of 3D Gaussians (right).} Each {\color{blue} blue} dot represents a projected 3D Gaussian center. Our proposal offers two key advantages: 1) Our 3D Gaussians are well-distributed across the scene, whereas CF-3DGS~\cite{Fu_2024_CVPR} has a notable absence of 3D Gaussians on the image's left side (e.g., in the {\color{red} red} region); 2) Our learned 3D Gaussians are of high quality. While CF-3DGS places numerous 3D Gaussians in the {\color{ForestGreen} green} region, the rendering quality there is notably inferior to ours.
    }
    \label{fig:teaser}
    \vspace{1.3em}
}]

\begin{abstract}
Standard 3D Gaussian Splatting (3DGS) relies on known or pre-computed camera poses and a sparse point cloud, obtained from structure-from-motion (SfM) preprocessing, to initialize and grow 3D Gaussians. We propose a novel SfM-Free 3DGS (SFGS) method for video input, eliminating the need for known camera poses and SfM preprocessing. Our approach introduces a hierarchical training strategy that trains and merges multiple 3D Gaussian representations -- each optimized for specific scene regions -- into a single, unified 3DGS model representing the entire scene. To compensate for large camera motions, we leverage video frame interpolation models. Additionally, we incorporate multi-source supervision to reduce overfitting and enhance representation. Experimental results reveal that our approach significantly surpasses state-of-the-art SfM-free novel view synthesis methods. On the Tanks and Temples dataset, we improve PSNR by an average of 2.25dB, with a maximum gain of 3.72dB in the best scene. On the CO3D-V2 dataset, we achieve an average PSNR boost of 1.74dB, with a top gain of 3.90dB. The code is available at \href{https://github.com/jibo27/3DGS_Hierarchical_Training/}{https://github.com/jibo27/3DGS\_Hierarchical\_Training}.
\end{abstract}
\vspace{-0.5cm}    
\section{Introduction}
\label{sec:intro}

3D Gaussian Splatting (3DGS)~\cite{kerbl20233d} represents a 3D scene from multi-view images based on camera intrinsic and extrinsic parameters along with an initial point cloud. 
Obtaining camera poses and the initial point cloud requires preprocessing, which is often performed using a structure-from-motion (SfM) algorithm~\cite{schonberger2016structure}. 
However, SfM can be time-consuming and may struggle with repetitive patterns, textureless regions, or feature extraction errors. Additionally, SfM lacks differentiability, which can limit its applicability in future research~\cite{bian2023nope}. 
As such, a class of new methods for novel view synthesis is trying to eliminate the need for SfM preprocessing~\cite{Fu_2024_CVPR,li2024ggrt,fan2024instantsplat,COGS2024}.

Removing SfM preprocessing introduces two obvious {questions} for 3DGS. 
{First, how can the camera poses of the input images be estimated? Second, how can 3D gaussians be initialized and grown within the scene?}
Inspired by CF-3DGS~\cite{Fu_2024_CVPR}, we address the challenge of constructing an SfM-free 3DGS from video sequences. Assuming video input with small camera movement, we address the issue of camera pose estimation by predicting the relative poses between temporally adjacent frames. By sequentially stacking these relative poses, we obtain the overall camera poses.

To improve camera pose estimation, a key innovation in our work is leveraging a video frame interpolation (VFI) model to generate additional frames. We using an off-the-shelf deep model~\cite{kong2022ifrnet} to double the input video length by interpolating between frames. 
Although these interpolated frames are not rendered from an underlying 3D model and may lack perfect geometric consistency, they provide sufficient quality (see Fig.~\ref{fig:vfi_help_pose_estimation}) to bridge relative poses between frames, which is particularly beneficial for sequences with larger camera movements. They also provide additional supervision, covering viewpoints not present in the original training frames. Incorporating these interpolated frames into 3DGS training yields a 0.35 dB performance boost on the Tanks and Temples dataset~\cite{Knapitsch2017}.

To address the second question of initializing and growing the 3D Gaussians, a straightforward way would be to use a point cloud derived from the depth map of the first frame; however, this often leads to sparse Gaussian coverage for regions not visible in the first frame.
The standard adaptive density control~\cite{kerbl20233d} -- which adjusts 3D Gaussians by splitting, cloning, and pruning -- struggles in these sparsely covered regions.  In these areas, the Gaussians may have very small gradients, making it challenging to activate densification processes~\cite{zhang2024pixel,bulo2024revising}. 

To this end, we propose a novel hierarchical training strategy that merges multiple base 3DGS models, each optimized for specific parts of the scene, into a unified model representing the entire scene. 
Intuitively, the adaptive density control encounters difficulties in regions with sparse 3D Gaussians; however, with our strategy, these regions are populated with 3D Gaussians merged from other 3DGS models. 
Interestingly, this merging strategy can be viewed as a densification process: we discard unimportant 3D Gaussians and densify the representation by merging essential Gaussians from different base 3DGS models. Fig~\ref{fig:teaser} illustrates the improved 3D Gaussian coverage achieved by our approach compared to a naive strategy without hierarchical training. This strategy boosts PSNR by 1.19–1.58 dB on the Tanks and Temples dataset\cite{Knapitsch2017}.

Furthermore, we enhance the representation quality through multi-source supervision, leveraging both base 3DGS models and interpolated frames from VFI. 

Our approach achieves a significant PSNR improvement of 2.25 dB on the Tanks and Temples~\cite{Knapitsch2017} and 1.74 dB on the CO3D-V2~\cite{reizenstein2021common} over state-of-the-art SfM-free novel view synthesis methods. Even without known camera intrinsics, our method surpasses the state-of-the-art methods by 0.89 dB in PSNR.
Our contributions are as follows:
\begin{itemize}
    \item We improve pose estimation by leveraging video frame interpolation to smooth camera motion.
    \item We introduce a hierarchical training strategy to address initialization and density control challenges without SfM preprocessing. Interestingly, this approach can be interpreted as a densification step.
    \item We employ multi-source supervision, reusing base 3DGS models and VFI-interpolated frames to reduce overfitting.
    \item Together, these innovations yield a 3DGS approach that requires no SfM preprocessing, significantly outperforming existing SfM-free novel view synthesis methods.
\end{itemize}

\section{Related Works}
\label{sec:related_works}

\textbf{Novel view synthesis} is the task of predicting realistic images from an unobserved viewpoint. NeRFs~\cite{mildenhall2021nerf} are an implicit 3D representation that encode scenes within an MLP. 
They are remarkable at rendering 
images from novel views, but despite the various 
proposed improvements~\cite{barron2021mip,barron2022mip,fridovich2022plenoxels,hu2023tri}, they are still relatively slow and may require up to several minutes to render a scene. 
In contrast, 3D Gaussian splats~\cite{kerbl20233d}, as an explicit representation, are able to achieve high-quality renderings at much faster speeds.  
Subsequent works build upon 3D Gaussian splatting, dealing with revising the density~\cite{zhang2024pixel,bulo2024revising}, compressing the representation~\cite{niedermayr2024compressed,lee2024compact,morgenstern2023compact} and anti-aliasing~\cite{yan2024multi,yu2024mip}. Others extend 3D Gaussian splatting to dynamic~\cite{duan20244d,shawswings,wu20244d} and large, city-scale
scenes~\cite{kerbl2024hierarchical,liu2024citygaussian}.\\

\noindent \textbf{SfM-free novel view synthesis} for both NeRFs and 3D Gaussian splatting is a class of works that try to do away with known or estimated camera pose from SfM. 
Examples include i-NeRF~\cite{yen2021inerf}, which estimates camera poses by aligning keypoints using a pre-trained NeRF. Follow-ups like 
NeRFmm~\cite{wang2021nerfmm}, SiNeRF~\cite{xia2022sinerf}, BARF~\cite{lin2021barf} and GARF~\cite{chng2022garf} learn both the NeRF model and camera pose embeddings simultaneously~\cite{wang2021nerfmm}, addressing the gradient inconsistency~\cite{lin2021barf,chng2022garf}, 
leveraging pre-trained networks for monocular depth estimation or optical flow, incorporating prior geometric knowledge or correspondence information~\cite{bian2023nope,cheng2023lu,meuleman2023progressively}. 

For 3D Gaussian splatting, CF-3DGS~\cite{Fu_2024_CVPR} and GGRt~\cite{li2024ggrt}, InstantSplat~\cite{fan2024instantsplat}, COGS~\cite{COGS2024} was developed to support SfM-free optimization.
CF-3DGS~\cite{Fu_2024_CVPR} performs an affine transformation on the positions of the 3D Gaussians to predict relative poses, progressively expanding the representations from the first frame to the last frame. However, its performance is limited by the accuracy of estimated camera poses.  It also suffers when there is insufficient initialization of 3D Gaussians, and has challenges in density control.
GGRt~\cite{li2024ggrt} jointly learns two modules for iterative pose optimization and a generalizable 3DGS. On the other hand, InstantSplat~\cite{fan2024instantsplat} and COGS~\cite{COGS2024} are designed primarily for scenarios with sparse image views. 
In this paper, we focus on the video input with small camera movement, similar to CF-3DGS~\cite{Fu_2024_CVPR}. We address two main challenges associated with applying 3DGS in SfM-free tasks: improving camera pose estimation and enhancing the initialization and learning of 3D Gaussians.

\begin{figure*}[t]
  \centering
   \includegraphics[width=0.95\linewidth]{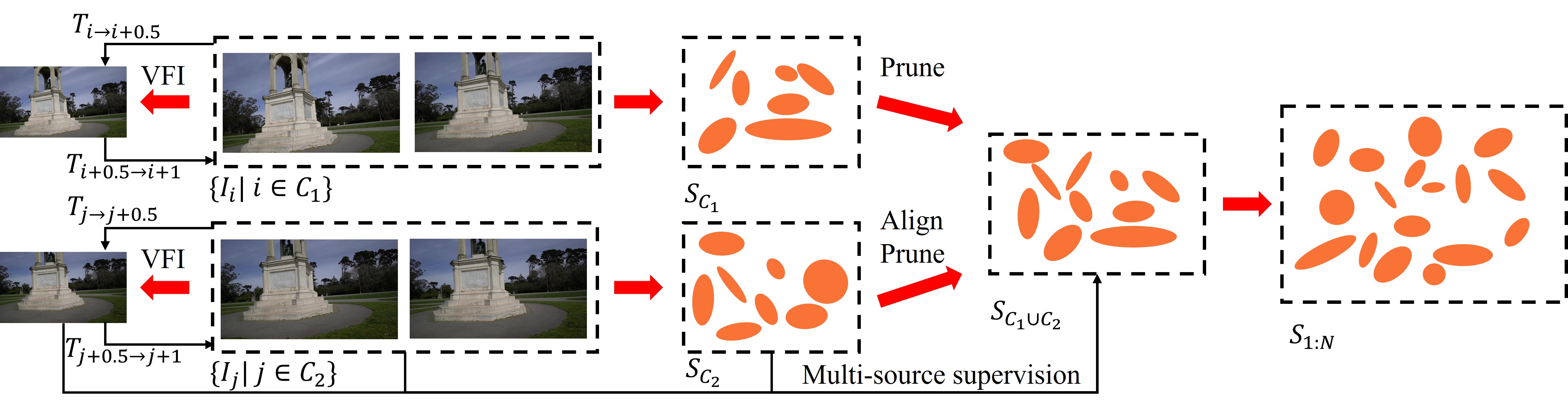}
   \vspace{-0.3cm}
   \caption{\textbf{Overview of our proposal.} We partition the video into multiple segments, train a base 3DGS model on each segment individually, and then iteratively merge these base models into a single, unified 3DGS model representing the entire scene.}
   \label{fig:overview}
   \vspace{-0.3cm}
\end{figure*}

\section{Approach}
\label{sec:approach}

\subsection{Overview}

Consider a video sequence $\mathcal{I} \!=\! \{I_i \mid i\!=\!1, \dots, N\}$ captured with small camera movements. 
We aim to reconstruct a 3D Gaussian splatting representation (3DGS) $\mathcal{S}$ from $\mathcal{I}$ and camera intrinsics $K$. 
We first estimate a series of camera poses $\mathcal{P}\!=\!\{P_i \mid i\!=\!1, \dots, N\}$  (Section~\ref{sec:camera_pose_estimation}).
Then, we partition the video into overlapping segments $\{C_j\}$. For each segment $C_j$, a base 3DGS model $\mathcal{S}_{C_j}$ is trained. These models are then iteratively merged from adjacent segments to form a unified representation (Section~\ref{sec:hierarchical_training}). After each merge, we retrain the merged 3DGS model using original training frames, pseudo-view frames from the base models, and interpolated frames from VFI on the combined segments (Section~\ref{sec:retraining}). This merging and retraining process continues until we obtain the final 3DGS model $\mathcal{S}$, representing the entire sequence $\mathcal{I}$. Fig.~\ref{fig:overview} provides an overview of the pipeline.

\subsection{Camera pose estimation}\label{sec:camera_pose_estimation}

We estimate the camera poses $\mathcal{P}$ by stacking relative camera poses between temporally adjacent pairs of frames. 
The camera pose of the first frame is set to have no rotation or translation, i.e., $P_1= [\mathbb{I}|\mathbf{0}]$, serving as the reference frame. The estimated poses for all subsequent frames are with respect to this first frame.
For each frame pair $(I_{i}, I_{i+1})$, we estimate the relative pose $T_{i \to i+1}$ across all $N-1$ pairs.
The camera pose for the $i$-th frame is the matrix multiplication of the previous relative camera poses:
\begin{align}
    P_{i}:= T_{1\to i} = T_{i-1 \to i}\odot \cdots \odot T_{2\to 3} \odot T_{1\to 2}.
\end{align}
While stacking relative poses can accumulate error, directly estimating each frame's pose with respect to the first frame is more challenging due to the larger camera displacement. \\

\noindent \textbf{Relative pose estimation.} 
As identified in~\cite{Fu_2024_CVPR}, the relative poses between two frames can be approximated by estimating an affine transformation, denoted as $A$, which is applied to 3D Gaussians from the first frame. After the transformation, the rendered image with respect to the second frame should 
align with the second frame.
Specifically, the 2D projection $\mu_{\text{2D}}$ of a 3D Gaussian with position $\mu$ under the pose $P$ is given by $\mu_{\text{2D}}\!=\!K (P\mu)/(P\mu)_z$. This can be approximated by applying an affine transformation $A$ to $\mu$, followed by a projection using the identity camera pose $[\mathbb{I}|0]$, yielding $\mu_{\text{2D}}\!=\!K (\mathbb{I}\mu')/(\mathbb{I}\mu')_z$, where $\mu'\!=\!A\mu$. As a result, the relative pose can be estimated from $A$.

In practice, to estimate the relative pose from $I_{i}$ to $I_{i+1}$, we first construct a single-image 3DGS model $\mathcal{S}_{i}$ optimized exclusively on $I_{i}$. 
We then apply the affine transformation $A$ to each Gaussian in $\mathcal{S}_{i}$ and render the image $\hat{I}_{i+1}$ using the camera pose $[\mathbb{I}|\mathbf{0}]$. 
The reconstructed $\hat{I}_{i+1}$ is expected to match $I_{i+1}$. We optimize $A$ by minimizing the photometric loss between the rendered image $\hat{I}_{i+1}$ and the target image $I_{i+1}$. 
During optimization, the attributes of 3D Gaussians in $\mathcal{S}_i$ are fixed, and only $A$ is optimized. 
The optimized transformation matrix $A$ corresponds to the estimated relative pose $T_{i \to i+1}$. \\

\begin{figure}[t]
    \centering
    \begin{subfigure}[t]{0.3\linewidth}
        \centering
        \includegraphics[width=\linewidth]{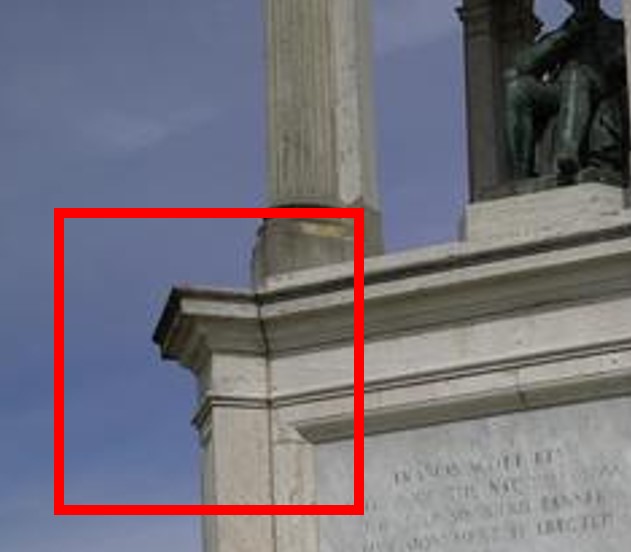}
        \caption{$I_{i}$}\label{fig:vfi_help_pose_estimation_I_prev}
    \end{subfigure}
    \begin{subfigure}[t]{0.3\linewidth}
        \centering
        \includegraphics[width=\linewidth]{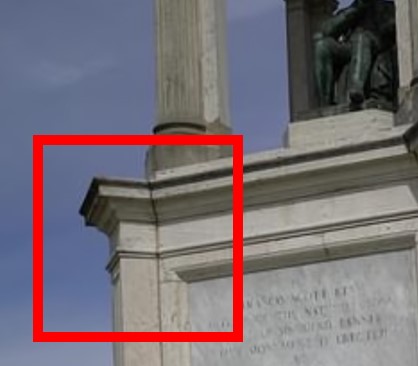}
        \caption{$I_{i+0.5}$}\label{fig:vfi_help_pose_estimation_I_interpolated}
    \end{subfigure}
    \begin{subfigure}[t]{0.3\linewidth}
        \centering
        \includegraphics[width=\linewidth]{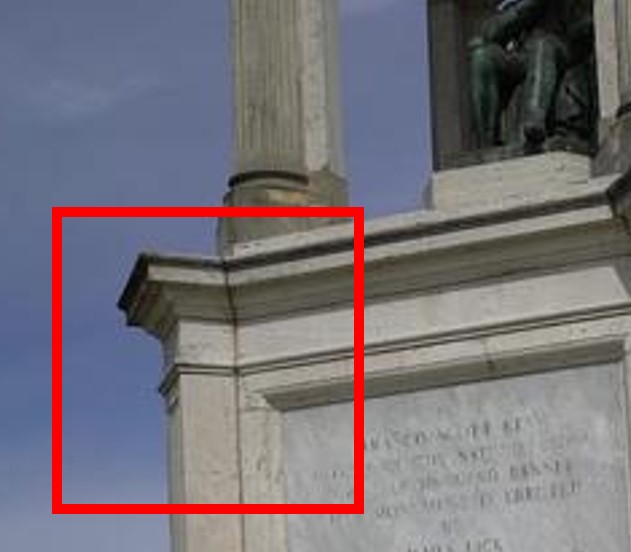}
        \caption{$I_{i+1}$}\label{fig:vfi_help_pose_estimation_I}
    \end{subfigure}
    \begin{subfigure}[t]{0.3\linewidth}
        \centering
        \includegraphics[width=\linewidth]{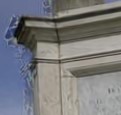}
        \caption{$\hat{I}_{i+1}$ w/o VFI}\label{fig:vfi_help_pose_estimation_I_wovfi}
    \end{subfigure}
    \begin{subfigure}[t]{0.3\linewidth}
        \centering
        \includegraphics[width=\linewidth]{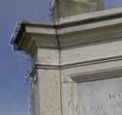}
        \caption{$\hat{I}_{i+0.5}$ w/ VFI}\label{fig:vfi_help_pose_estimation_I_05_withvfi}
    \end{subfigure}
    \begin{subfigure}[t]{0.3\linewidth}
        \centering
        \includegraphics[width=\linewidth]{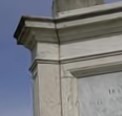}
        \caption{$\hat{I}_{i+1}$ w/ VFI}\label{fig:vfi_help_pose_estimation_I_withvfi}
    \end{subfigure}
    \vspace{-0.2cm}
    \caption{\textbf{Effect of VFI on relative pose estimation between $I_{i}$  (\ref{fig:vfi_help_pose_estimation_I_prev}) and $I_{i+1}$ (\ref{fig:vfi_help_pose_estimation_I}).} Fig~\ref{fig:vfi_help_pose_estimation_I_interpolated} shows the interpolated frame.
    In Fig~\ref{fig:vfi_help_pose_estimation_I_wovfi}, artifacts are noticeable in regions affected by camera movement, which VFI helps reduce. Fig~\ref{fig:vfi_help_pose_estimation_I_05_withvfi} and \ref{fig:vfi_help_pose_estimation_I_withvfi} show fewer artifacts in the rendered interpolated and original frames.
    }
    \label{fig:vfi_help_pose_estimation}
    \vspace{-0.5cm}
\end{figure}

\noindent \textbf{Relative pose estimation with video frame interpolation.} 
When the camera movement between adjacent frames is small, the relative pose estimation described above performs well because of sufficient frame overlap. However, with larger camera motions, performance decreases, and the fixed single-image 3DGS model from the previous frame may fail to render a high-quality image $\hat{I}_{i+1}$. 
This is because larger camera motions introduce more unseen content that the previous frame may not cover, leading to optimization objectives impacted by artifacts and resulting in poorly estimated poses. For example, the rendered $\hat{I}_{i+1}$ exhibits artifacts in such regions, as shown in Fig.~\ref{fig:vfi_help_pose_estimation_I_wovfi}.

A key insight of our work is compensating for large camera motions with a well-trained video frame interpolation (VFI) model~\cite{kong2022ifrnet}. 
Let $I_{i+0.5}$ represent an interpolated frame between $I_{i}$ and $I_{i+1}$, where the decimal $0.5$ indicates the interpolated result. We estimate the relative poses between $I_{i}$ and $I_{i+0.5}$, and between $I_{i+0.5}$ and $I_{i+1}$, separately. The overall relative pose is then given by $T_{i\to i+1} \!=\! T_{i\to i+0.5} \odot T_{i+0.5 \to i+1}$.
By reducing the relative camera motion in each step, fewer artifacts are introduced (see Figs.~\ref{fig:vfi_help_pose_estimation_I_05_withvfi} and \ref{fig:vfi_help_pose_estimation_I_withvfi}).

\subsection{Hierarchical Training }\label{sec:hierarchical_training}

After estimating the camera poses $\mathcal{P}$, the next step is to initialize and grow 3D Gaussians from the input video. A straightforward approach is to initialize the 3D Gaussians using point clouds lifted from the depth map~\cite{ranftl2021vision, bhat2023zoedepth} of the first frame, then gradually grow the 3D Gaussians by processing frames sequentially. 
This initialization, based on a single frame, is incomplete in covering the scene. 
Moreover, the standard {adaptive density control}~\cite{kerbl20233d}, which relies on accumulated gradients to split and clone Gaussians or prunes them based on low opacity, struggles in regions with sparse Gaussians distributions. In such areas, Gaussians may have very small gradients, and thus fail to activate the densification process. Consequently, as shown in Fig.~\ref{fig:teaser}, areas in the scene not covered in the first frame exhibit a noticeable lack of Gaussians. To overcome this limitation, we propose the hierarchical training strategy. \\

\noindent \textbf{Video partitioning.} 
We first partition the $N$ frames into overlapping segments, with the $j$-th segment denoted as $C_j$.
Ideally, each segment features similar scene content so that the 3DGS model is trained without encountering many unseen regions. 
To achieve this, we reuse the estimated 
relative camera poses. 
Segments $C_j$ and $C_{j+1}$ are separated by locating adjacent image pairs with the largest camera movement.  
Empirically, we observe that evenly partitioning the video into segments yields similar results. \\ 

\noindent \textbf{Training of base 3DGS models.} 
After partitioning the video into segments, we train the 3DGS model $\mathcal{S}_{C_j}$ for each segment $C_j$. The 3D Gaussians are initialized using a point cloud predicted from the first frame of $C_j$ by depth estimators~\cite{ranftl2021vision,bhat2023zoedepth}. The Gaussians are then optimized from the first to the last frame in each segment, growing as needed to capture the scene details. \\

\noindent \textbf{Merging of base 3DGS models.}
To represent the entire scene, we merge these base 3DGS models.
Consider the merging of two base 3DGS models $\mathcal{S}_{C_j}$ and $\mathcal{S}_{C_{j+1}}$. Since each model is optimized with its 3D Gaussian positions aligned to the first frame of its respective segment, it is necessary to align the two models before merging. The alignment is performed by transforming the 3D Gaussians in $\mathcal{S}_{C_{j+1}}$ to match those in $\mathcal{S}_{C_j}$, based on the relative pose $T_{C_{j+1,1} \to C_{j,1}}$, where $C_{j,1}$ and $C_{j+1,1}$ denote the first frames of their respective segment.

For merging two 3DGS models, one potential idea is to identify correspondences between 3D Gaussians in the two models and interpolate matched pairs. However, this idea presents several challenges in our task. First, establishing accurate correspondences between Gaussians is non-trivial.
For instance, one should not interpolate the Gaussians in the non-overlapping regions in the scene. Yet distinguishing between overlapping and non-overlapping regions can be difficult.
Secondly, even with correspondences identified, variations in properties like opacity, scale, covariance, and color among matched Gaussians could complicate the interpolation process~\cite{kerbl2024hierarchical}.

Therefore, instead of interpolation, we take a simple yet effective merging strategy: first, prune unimportant 3D Gaussians in each model, then concatenate the remaining Gaussians. 
Specifically, we first assign an importance score to each 3D Gaussian.
Our importance score is inspired by a 3DGS compression strategy~\cite{niedermayr2024compressed}, which defines the importance of a given parameter $p$ for a 3D Gaussian as:
\begin{align}
    f(p) = \frac{1}{\sum_{i=1}^N H_i W_i } \sum_{i=1}^N \left|\frac{\partial \left(\sum_{x,y}\hat{I}_i(x,y)\right)}{\partial p}\right|.\label{eq:importance_score}
\end{align}
Above, $\sum_{x,y}\hat{I}_i(x,y)$ represents the sum of RGB values in the rendered image $\hat{I}_i$, and $H_i$ and $W_i$ are the height and width of image $\hat{I}_i$. Parameter $p$ is a general notation representing variables such as color $c$, opacity $\alpha$, or covariance.
The importance score evaluates the sensitivity of the rendering quality with respect to changes to parameter $p$ of a 3D Gaussian. If a minor change in the parameter significantly affects the rendered image, that parameter is considered important.
We compute the importance score for each 3D Gaussian in base 3DGS models using Eq~\ref{eq:importance_score} and 
keep the top $\gamma$ percent from base 3DGS models. Finally, we take the union of the selected 3D Gaussians to form the merged representation.

It is feasible to take the union since 3DGS is an explicit representation. While this approach may introduce redundant Gaussians, we empirically observe that adaptive density control~\cite{kerbl20233d} is more effective at pruning redundant Gaussians than at generating new or high-quality Gaussians in sparse regions. 
If we anchor the model to $\mathcal{S}_{C_j}$, our merging strategy can be viewed as a densification step: it removes less important Gaussians and densifies the representation by adding importance Gaussians from $\mathcal{S}_{C_{j+1}}$. \\

\noindent \textbf{Hierarchical training.} 
After defining the partition and merging strategy, we describe the hierarchical training pipeline that merges multiple base 3DGS models, each optimized for individual segments, into a unified model. First, we define a hierarchical level $L$ and partition the input into $2^L$ overlapping segments, resulting in $2^L$ base models. We iteratively merge adjacent pairs, reducing the number of models by half in each step, until only one unified model remains.
For example, with $L\!=\!2$, we create four base models: $\mathcal{S}_{C_1}$, $\mathcal{S}_{C_2}$, $\mathcal{S}_{C_3}$, and $\mathcal{S}_{C_4}$. The first merge yields $\mathcal{S}_{C_1 \cup C_2}$ and $\mathcal{S}_{C_3 \cup C_4}$, and the final merge gives $\mathcal{S}_{1:N} \!:=\! \mathcal{S}_{C_1 \cup C_2 \cup C_3 \cup C_4}$.
Hierarchical training requires all input frames in advance. We also explore an online approach in which we sequentially merge $\mathcal{S}_{C_1 \cup C_2}$ with $\mathcal{S}_{C_3}$, then with $\mathcal{S}_{C_4}$, yielding $\mathcal{S}_{1:N}$. We refer to this variant as \textit{progressive training}. Both strategies improve the baseline by at least 1.32dB (Table~\ref{tab:abs_overview}), with slightly better results for hierarchical training.

\subsection{Multi-source supervision}\label{sec:retraining}

After merging $\mathcal{S}_{C_j}$ and $\mathcal{S}_{C_{j+1}}$ 
and pruning unimportant 3D Gaussians, the newly merged representation needs to be retrained. 
Simply retraining on the set of images from $C_j \cup C_{j+1}$ leads to overfitting on those specific images. To address this, we propose to augment training with two additional sets of images: (1) from the base 3DGS models $\mathcal{S}_{C_j}$ and $\mathcal{S}_{C_{j+1}}$ before merging, and (2) from the interpolated frames (see Section~\ref{sec:camera_pose_estimation}). \\

\noindent \textbf{Supervision from base 3DGS models.}
As the base 3DGS models are better-optimized for their respective segments, novel pseudo-views rendered from these models can serve as a source of training images for the merged 3DGS.
To generate novel views, we first sample a virtual camera pose $P_{i+\tau}$ between two poses, $P_{i}$ and $P_{i+1}$, using the formula:
\begin{align}
    P_{i+\tau} = P_{i} \exp \left(\tau \log \left(P_{i}^{-1} \cdot P_{i+1}\right)\right),
\end{align}
where $\tau \in (0, 1)$ and $P_{i}, P_{i+1} \in \text{SE}(3)$ represent camera poses in the $\text{SE}(3)$ space.
This smooth interpolation enables the creation of pseudo-views, which are then rendered as additional supervision for the merged 3DGS model. \\

\noindent \textbf{Supervision from video frame interpolation.}
The interpolated images from VFI are of sufficiently high quality at viewpoints not covered by the training frames, making them suitable for supervision.
For instance, Fig.~\ref{fig:vfi_help_pose_estimation_I_interpolated} shows an example of an interpolated frame.
To supervise the merged 3DGS with the interpolated frames, we need to estimate their corresponding camera pose.  
For an interpolated frame $I_{i+0.5}$ between frames $I_{i}$ and $I_{i+1}$, the camera pose $P_{i+0.5}$ is computed as $P_{i+0.5} = T_{i \to i+0.5} \odot \cdots \odot T_{1.5 \to 2} \odot T_{1 \to 1.5}$. Each relative pose was previously calculated in Section~\ref{sec:camera_pose_estimation}, so no additional computational overhead is required. \\

\noindent\textbf{Loss function.}
We optimize the 3DGS model using the photometric loss between the rendered image and the training image or the pseudo image~\cite{kerbl20233d}:
\begin{align}
    \mathcal{L} = (1 - \lambda) \mathcal{L}_1  + \lambda \mathcal{L}_{\text{D-SSIM}},
\end{align}
where $\mathcal{L}_1$ is the L1 loss, and $\mathcal{L}_{\text{D-SSIM}}$ is the D-SSIM term.

\section{Experiment}
\label{sec:experiment}

\begin{table*}[ht]
  \centering
  \resizebox{\textwidth}{!}{%
  \begin{tabular}{c|ccc|ccc|ccc|ccc|ccc}
    \toprule
     \multirow{2}{*}{Scenes}   & \multicolumn{3}{c|}{BARF~\cite{lin2021barf}}  & \multicolumn{3}{c|}{SC-NeRF~\cite{jeong2021self}} &  \multicolumn{3}{c|}{Nope-NeRF~\cite{bian2023nope}}  &  \multicolumn{3}{c|}{CF-3DGS~\cite{Fu_2024_CVPR}} &  \multicolumn{3}{c}{Ours}\\
    & PSNR $\uparrow$ & SSIM $\uparrow$ & LPIPS $\downarrow$ & PSNR $\uparrow$ & SSIM $\uparrow$ & LPIPS $\downarrow$  & PSNR $\uparrow$ & SSIM $\uparrow$ & LPIPS $\downarrow$  & PSNR $\uparrow$ & SSIM $\uparrow$ & LPIPS $\downarrow$ & PSNR $\uparrow$ & SSIM $\uparrow$ & LPIPS $\downarrow$  \\
    \midrule
     Church   & 23.17 & 0.62 & 0.52 & 21.96 & 0.60 & 0.53 & 25.17 & 0.73 & 0.39 & 30.23 & 0.93 & 0.11 & \cellcolor{bgcolor}31.34 & \cellcolor{bgcolor}0.94 & \cellcolor{bgcolor}0.08 \\ 
     Barn     & 25.28 & 0.64 & 0.48 & 23.26 & 0.62 & 0.51 & 26.35 & 0.69 & 0.44 & 31.23 & 0.90 & 0.10 & \cellcolor{bgcolor}34.95 & \cellcolor{bgcolor}0.97 & \cellcolor{bgcolor}0.05 \\
     Museum   & 23.58 & 0.61 & 0.55 & 24.94 & 0.69 & 0.45 & 26.77 & 0.76 & 0.35 & 29.91 & 0.91 & 0.11 & \cellcolor{bgcolor}31.59 & \cellcolor{bgcolor}0.95 & \cellcolor{bgcolor}0.08 \\
     Family   & 23.04 & 0.61 & 0.56 & 22.60 & 0.63 & 0.51 & 26.01 & 0.74 & 0.41 & 31.27 & 0.94 & 0.07 & \cellcolor{bgcolor}34.71 & \cellcolor{bgcolor}0.97 & \cellcolor{bgcolor}0.05\\
     Horse    & 24.09 & 0.72 & 0.41 & 25.23 & 0.76 & 0.37 & 27.64 & 0.84 & 0.26 & 33.94 & 0.96 & 0.05 & \cellcolor{bgcolor}35.82 & \cellcolor{bgcolor}0.98 & \cellcolor{bgcolor}0.03 \\
     Ballroom & 20.66 & 0.50 & 0.60 & 22.64 & 0.61 & 0.48 & 25.33 & 0.72 & 0.38 & 32.47 & 0.96 & 0.07 & \cellcolor{bgcolor}34.12 & \cellcolor{bgcolor}0.97 & \cellcolor{bgcolor}0.04 \\
     Francis  & 25.85 & 0.69 & 0.57 & 26.46 & 0.73 & 0.49 & 29.48 & 0.80 & 0.38 & 32.72 & 0.91 & 0.14 & \cellcolor{bgcolor}34.09 & \cellcolor{bgcolor}0.93 & \cellcolor{bgcolor}0.13 \\
     Ignatius & 21.78 & 0.47 & 0.60 & 23.00 & 0.55 & 0.53 & 23.96 & 0.61 & 0.47 & 28.43 & 0.90 & 0.09 & \cellcolor{bgcolor}31.64 & \cellcolor{bgcolor}0.95 & \cellcolor{bgcolor}0.06  \\ 
    \midrule
     Mean     & 23.42 & 0.61 & 0.54 & 23.76 & 0.65 & 0.48 & 26.34 & 0.74 & 0.39 & 31.28 & 0.93 & 0.09 & \cellcolor{bgcolor}33.53 & \cellcolor{bgcolor}0.96 & \cellcolor{bgcolor}0.07 \\ 
    \bottomrule
  \end{tabular}
  }
  \caption{\textbf{Novel view synthesis results on Tanks and Temples~\cite{Knapitsch2017}.} We achieve the best performance among all competitors. 
  }
  \label{tab:sota_nvs_tanks}
\end{table*}

\begin{figure*}[t]
    \centering
    \begin{subfigure}[t]{\linewidth}
        \centering
        \includegraphics[width=\linewidth]{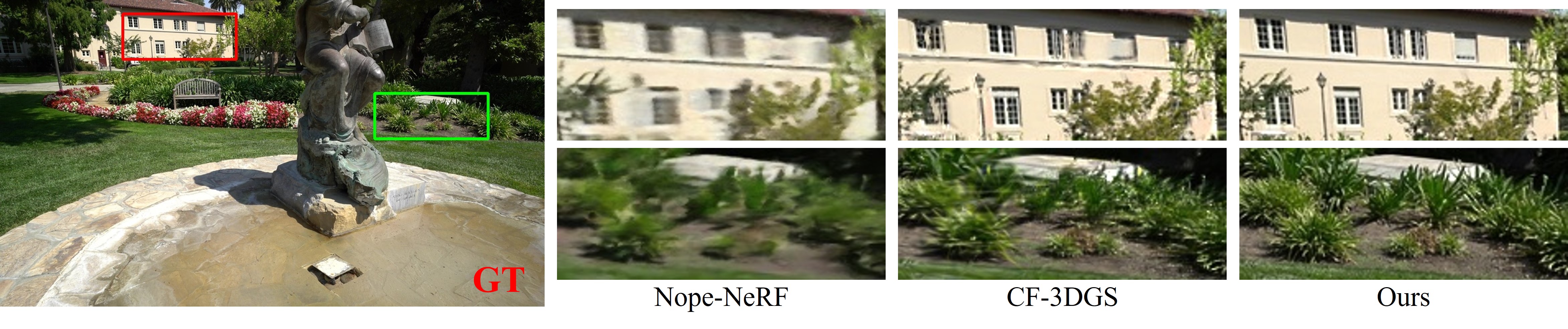}
    \end{subfigure}
    \begin{subfigure}[t]{\linewidth}
        \centering
        \includegraphics[width=\linewidth]{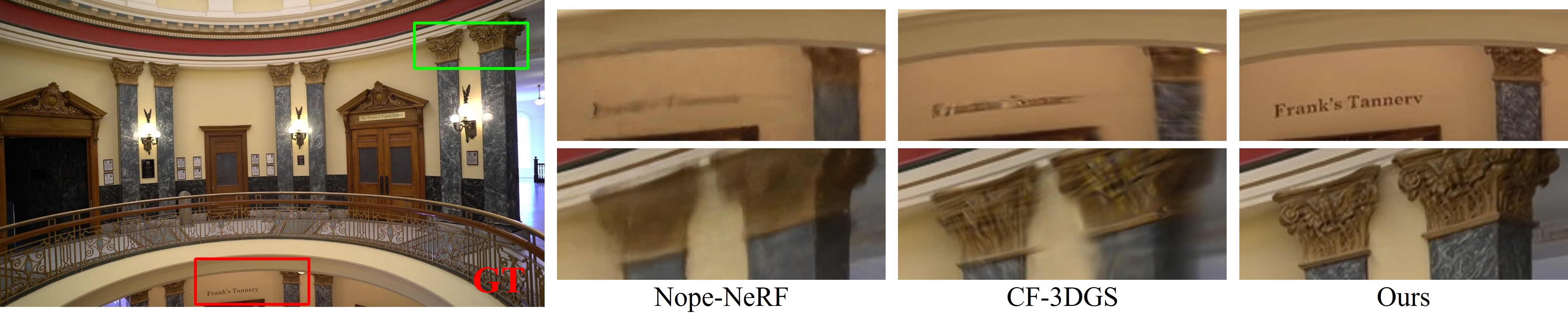}
    \end{subfigure}
    \vspace{-0.5cm}
    \caption{\textbf{Qualitative novel view synthesis results on Tanks and Temples~\cite{Knapitsch2017}.} Our proposal achieves superior rendering quality. 
    }
    \label{fig:sota_nvs_tanks}
    \vspace{-0.4cm}
\end{figure*}

\subsection{Experimental setting}
\noindent \textbf{Datasets.} 
We conduct experiments on the Tanks and Temples dataset~\cite{Knapitsch2017} and the CO3D-V2 dataset~\cite{reizenstein2021common}. For Tanks and Temples, following~\cite{bian2023nope, Fu_2024_CVPR}, we sequentially divide the frames into groups of 8, using one of the frames as the test frame and the remaining frames for training. For the \textit{Family} scene on Tanks and Temples, we alternate images, using every other image as a test image. 
The CO3D-V2 dataset is more challenging due to its larger camera motions. We adopt the same sampling strategy, using every eighth frame for testing and the rest for training. 

\noindent \textbf{Metrics.}
We use PSNR, SSIM~\cite{wang2004image}, and LPIPS~\cite{zhang2018unreasonable} to evaluate the effectiveness of novel view synthesis. For camera pose estimation, we report the Absolute Trajectory Error (ATE) and Relative Pose Error (RPE), similar to ~\cite{bian2023nope,Fu_2024_CVPR}. ATE measures the difference of the camera positions. RPE measures the relative pose errors, containing relative rotation error ($\text{RPE}_r$) and relative translation error ($\text{RPE}_t$). Additionally, we report the memory size required to store the optimized parameters.

\noindent \textbf{Implementation details.} 
The hierarchical training level is set to $L\!=\!2$ as it reaches the saturation point (see Table~\ref{tab:abs_hierarchical_level}).
Each segment's base 3DGS model is trained from start to end frame with 300 iterations per frame. Multi-source supervision also employs 300 iterations per frame. During the merging, we select the top 50\% of 3D Gaussians from each model based on importance score.
Multi-source supervision involves two steps: first, using pseudo-view images from base models and original frames; second, using interpolated frames and original frames, with a 50\% probability of selecting pseudo-view or interpolated frame for each step. Gaussians are grown and pruned every 100 and 2000 iterations on the Tanks and Temples and CO3D-V2 datasets, respectively. 
All experiments are conducted on a single RTX A5000 GPU.
More details are in the supplementary material.

\begin{table*}[ht]
  \centering
  \resizebox{\textwidth}{!}{
  \begin{tabular}{c|ccc|ccc|ccc|ccc|ccc}
    \toprule
     \multirow{2}{*}{Method}  &  \multicolumn{3}{c|}{34\_1403\_4393} & \multicolumn{3}{c|}{106\_12648\_23157} & \multicolumn{3}{c|}{110\_13051\_23361} & \multicolumn{3}{c|}{245\_26182\_52130} & \multicolumn{3}{c}{415\_57112\_110099} \\
    & PSNR $\uparrow$ & SSIM $\uparrow$ & LPIPS $\downarrow$ & PSNR $\uparrow$ & SSIM $\uparrow$ & LPIPS $\downarrow$ & PSNR $\uparrow$ & SSIM $\uparrow$ & LPIPS $\downarrow$ & PSNR $\uparrow$ & SSIM $\uparrow$ & LPIPS $\downarrow$  & PSNR $\uparrow$ & SSIM $\uparrow$ & LPIPS $\downarrow$  \\
    \midrule
    Nope-NeRF~\cite{bian2023nope} & 28.62 & 0.80 & 0.35 & 20.41 & 0.46 & 0.58 & 26.86 & 0.73 & 0.47 & 25.05 & 0.80 & 0.49 & 24.78 & 0.64 & 0.55 \\
    CF-3DGS~\cite{Fu_2024_CVPR}   & 27.75 & 0.86 & 0.20 & 22.14 & 0.64 & 0.34 & 29.69 & \cellcolor{bgcolor}0.89 & 0.29 & 27.24 & 0.85 & 0.30 & 26.21 & 0.73 & 0.32 \\
    Ours                          & \cellcolor{bgcolor}32.52 & \cellcolor{bgcolor}0.93 & \cellcolor{bgcolor}0.14 & \cellcolor{bgcolor}23.43 & \cellcolor{bgcolor}0.73 & \cellcolor{bgcolor}0.28 & \cellcolor{bgcolor}29.95 & 0.87 & \cellcolor{bgcolor}0.19 & \cellcolor{bgcolor}28.59 & \cellcolor{bgcolor}0.87 & \cellcolor{bgcolor}0.27 & \cellcolor{bgcolor}27.23 & \cellcolor{bgcolor}0.78 & \cellcolor{bgcolor}0.30 \\
    \bottomrule
  \end{tabular}
  }
  \caption{\textbf{Novel view synthesis results on CO3D-V2~\cite{reizenstein2021common}.}  We achieve superior performance over all competitors, with the largest improvement on \textit{34\_1403\_4393}, where our method increases PSNR by 3.90 dB compared to Nope-NeRF.
  }
  \label{tab:sota_nvs_co3d}
\end{table*}

\begin{figure*}[t]
    \centering
    \begin{subfigure}[t]{\linewidth}
        \centering
        \includegraphics[width=\linewidth]{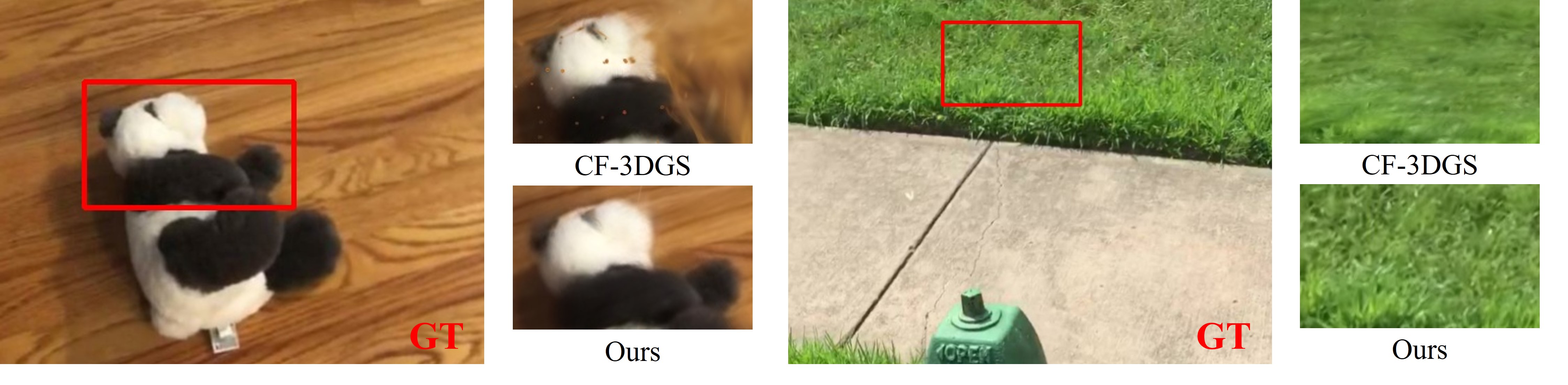}
    \end{subfigure}
    \begin{subfigure}[t]{\linewidth}
        \centering
        \includegraphics[width=\linewidth]{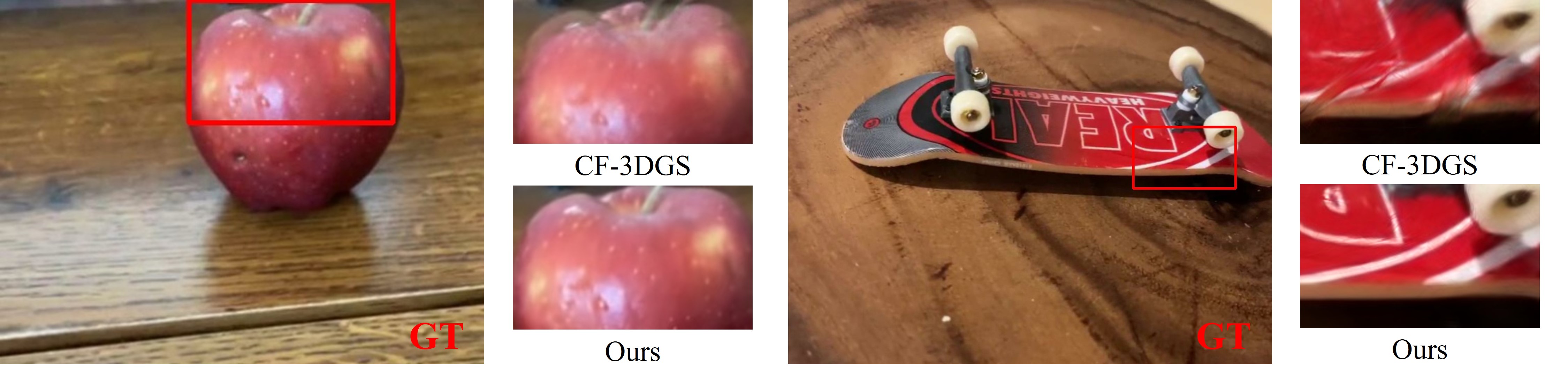}
    \end{subfigure}
    \vspace{-0.5cm}
    \caption{\textbf{Qualitative novel view synthesis results on CO3D-V2~\cite{reizenstein2021common}.} Our proposal achieves superior rendering quality. 
    }
    \label{fig:sota_nvs_co3d}
    \vspace{-0.3cm}
\end{figure*}

\subsection{Evaluation on the Tanks and Temples Dataset}
\noindent \textbf{Quantitative comparison.}
We perform the comparison with the state-of-the-art novel view synthesis methods without SfM preprocessing, including BARF~\cite{lin2021barf}, SC-NeRF~\cite{jeong2021self}, Nope-NeRF~\cite{bian2023nope} and CF-3DGS~\cite{Fu_2024_CVPR} on Tanks and Temples~\cite{Knapitsch2017}. 
As shown in Table~\ref{tab:sota_nvs_tanks}, our method achieves superior performance compared to all of them. Specifically, compared to CF-3DGS, we improve the average PSNR by 2.25 dB, SSIM by 0.03, and reduce LPIPS by 0.02. The most significant improvement is observed in the \textit{Barn} scene, with a PSNR increase of 3.72 dB, SSIM improvement of 0.07, and LPIPS reduction of 0.05.

\noindent \textbf{Qualitative comparison.} Fig.~\ref{fig:sota_nvs_tanks} shows that we achieve finer detail and higher fidelity, especially in highly detailed regions where CF-3DGS struggles to grow 3D Gaussians, highlighting the advantages of our training strategy.

\subsection{Evaluation on the CO3D-V2 Dataset}
\textbf{Quantitative comparison.}
Due to the challenges posed by this dataset, we limit our comparison to the most advanced methods, namely Nope-NeRF~\cite{bian2023nope} and CF-3DGS~\cite{Fu_2024_CVPR}. 
Table~\ref{tab:sota_nvs_co3d} shows that our method outperforms these approaches, with an average PSNR boost of 1.74 dB. 
The most significant improvement is seen in the scene \textit{34\_1403\_4393} (teddybear), where we increase the PSNR by over 3.90 dB, improve SSIM by 0.07, and reduce LPIPS by 0.06. 

\noindent \textbf{Qualitative comparison.} 
Fig.~\ref{fig:sota_nvs_co3d} demonstrates that even with challenging input videos, our method maintains high performance compared to CF-3DGS, which exhibits blur and unrealistic red artifacts in the \textit{34\_1403\_4393} (teddybear) scene due to suboptimal 3D Gaussian learning.

\noindent \textbf{Camera pose estimation.}
We conduct camera pose estimation comparisons only on CO3D-V2, as it provides ground-truth camera poses. As shown in Table~\ref{tab:sota_pose_co3d}, our method achieves comparable or superior performance to all competitors, reducing $\text{RPE}_t$ and $\text{RPE}_r$ by up to 0.464 and 0.078, respectively. 
Our approach's improvements on $\text{ATE}$ are less consistent. We hypothesize that this is due to errors in the interpolated images generated by the VFI model, to which camera position estimation is particularly sensitive.

\begin{table*}[ht]
  \centering
  \resizebox{\textwidth}{!}{
  \begin{tabular}{c|ccc|ccc|ccc|ccc|ccc}
    \toprule
    \multirow{2}{*}{Method}  &  \multicolumn{3}{c|}{34\_1403\_4393} &  \multicolumn{3}{c|}{106\_12648\_23157} & \multicolumn{3}{c|}{110\_13051\_23361} & \multicolumn{3}{c|}{245\_26182\_52130}  & \multicolumn{3}{c}{415\_57112\_110099} \\
    & $\text{RPE}_t$ $\downarrow$ & $\text{RPE}_r$ $\downarrow$ & $\text{ATE}$ $\downarrow$ & $\text{RPE}_t$ $\downarrow$ & $\text{RPE}_r$ $\downarrow$ & $\text{ATE}$ $\downarrow$  & $\text{RPE}_t$ $\downarrow$ & $\text{RPE}_r$ $\downarrow$ & $\text{ATE}$ $\downarrow$  & $\text{RPE}_t$ $\downarrow$ & $\text{RPE}_r$ $\downarrow$ & $\text{ATE}$ $\downarrow$  & $\text{RPE}_t$ $\downarrow$ & $\text{RPE}_r$ $\downarrow$ & $\text{ATE}$ $\downarrow$ \\
    \midrule
    Nope-NeRF~\cite{bian2023nope} & 0.591      & 1.313 & 0.053 & 0.387 & 1.312 & 0.049 & 0.400 & 1.966 & 0.046 & 0.587 & 1.867 & 0.038 & 0.326 & 1.919 & 0.054\\
    CF-3DGS~\cite{Fu_2024_CVPR}   & 0.505      & 0.211 & \cellcolor{bgcolor}0.009 & 0.094 & 0.360 & \cellcolor{bgcolor}0.008 & 0.140 & 0.401 & 0.021 & 0.239 & 0.472 & \cellcolor{bgcolor}0.017 & 0.110 & 0.424 & \cellcolor{bgcolor}0.014\\
    Ours                          & \cellcolor{bgcolor}0.041 & \cellcolor{bgcolor}0.170 & \cellcolor{bgcolor}0.009 & \cellcolor{bgcolor}0.045 & \cellcolor{bgcolor}0.282 & 0.014 & \cellcolor{bgcolor}0.093 & \cellcolor{bgcolor}0.331 & \cellcolor{bgcolor}0.020 & \cellcolor{bgcolor}0.064 & \cellcolor{bgcolor}0.438 & \cellcolor{bgcolor}0.017 & \cellcolor{bgcolor}0.049 &    \cellcolor{bgcolor}0.351 & 0.024\\
    \bottomrule
  \end{tabular}
  }
  \caption{\textbf{Camera pose estimation results on CO3D-V2~\cite{reizenstein2021common}.} We achieve significant improvements in $\text{RPE}_t$ and $\text{RPE}_r$, with a slight decrease in performance on $\text{ATE}$. We hypothesize that $\text{ATE}$ is more sensitive to errors in the interpolated frames from VFI.}
  \label{tab:sota_pose_co3d}
\end{table*}

\begin{table*}[ht]
  \centering
  \resizebox{\textwidth}{!}{%
  \begin{tabular}{c|cccc|cccc|cccc|cccc}
    \toprule
    \multirow{2}{*}{Scenes}   & \multicolumn{4}{c|}{$L=0$} &  \multicolumn{4}{c|}{$L=1$}  &  \multicolumn{4}{c|}{$L=2$} &  \multicolumn{4}{c}{$L=3$}\\
    & PSNR $\uparrow$ & SSIM $\uparrow$ & LPIPS $\downarrow$ & Mem $\downarrow$ & PSNR $\uparrow$ & SSIM $\uparrow$ & LPIPS $\downarrow$ & Mem $\downarrow$ & PSNR $\uparrow$ & SSIM $\uparrow$ & LPIPS $\downarrow$ & Mem $\downarrow$ & PSNR $\uparrow$ & SSIM $\uparrow$ & LPIPS $\downarrow$ & Mem $\downarrow$ \\
    \midrule
     Church   & 30.44 & 0.93 & 0.09 & 1.06 & 31.40 & 0.94 & 0.08 & 0.84 & \cellcolor{bgcolor}31.67 & 0.95 & 0.08 & 0.85 & 31.61 & 0.94 & 0.08 & 0.83  \\
     Barn     & 30.09 & 0.88 & 0.11 & 1.44 & 32.06 & 0.92 & 0.08 & 1.43 & \cellcolor{bgcolor}32.27 & 0.92 & 0.08 & 1.41 & 32.20 & 0.92 & 0.08 & 1.38   \\
     Museum   & 30.24 & 0.91 & 0.10 & 1.35 & 30.95 & 0.94 & 0.08 & 1.17 & 31.75 & 0.94 & 0.07 & 1.10 & \cellcolor{bgcolor}31.97 & 0.95 & 0.07 & 1.09   \\
     Family   & 33.12 & 0.96 & 0.05 & 1.44 & 33.71 & 0.96 & 0.05 & 1.18 & \cellcolor{bgcolor}34.20 & 0.97 & 0.05 & 1.21 & 34.15 & 0.97 & 0.05 & 1.18   \\
     Horse    & 34.08 & 0.96 & 0.05 & 1.07 & 35.44 & 0.98 & 0.04 & 0.96 & 35.44 & 0.98 & 0.04 & 0.90 & \cellcolor{bgcolor}35.54 & 0.98 & 0.03 & 0.92 \\
     Ballroom & 32.82 & 0.96 & 0.05 & 1.39 & 33.17 & 0.97 & 0.05 & 1.14 & 33.41 & 0.97 & 0.04 & 1.14 & \cellcolor{bgcolor}33.67 & 0.97 & 0.04 & 1.13 \\
     Francis  & 32.84 & 0.92 & 0.14 & 0.81 & 33.64 & 0.92 & 0.13 & 0.68 & \cellcolor{bgcolor}33.66 & 0.92 & 0.13 & 0.75 & 33.62 & 0.92 & 0.13 & 0.64  \\
     Ignatius & 28.37 & 0.91 & 0.09 & 2.09 & 31.15 & 0.94 & 0.06 & 1.50 & 31.78 & 0.94 & 0.06 & 1.43 & \cellcolor{bgcolor}31.90 & 0.95 & 0.06 & 1.40   \\
    \midrule
     Mean     & 31.50 & 0.93 & 0.09 & 1.33 & 32.69 & 0.95 & 0.08 & 1.11 & 33.02 & 0.95 & 0.07 & 1.10 & \cellcolor{bgcolor}33.08 & 0.95 & 0.07 & 1.07  \\
    \bottomrule
  \end{tabular}
  }
  \caption{\textbf{Ablation study of the hierarchical training level.} 
  Memory storage is measured in gigabytes.}
  \label{tab:abs_hierarchical_level}
  \vspace{-0.2cm}
\end{table*}

\begin{table}[ht]
  \centering
  \resizebox{\linewidth}{!}{
  \begin{tabular}{clccc}
    \toprule
    Id & Variant & PSNR $\uparrow$ & SSIM $\uparrow$ & LPIPS $\downarrow$  \\
    \midrule
    (1) & CF-3DGS & 31.28 & 0.93 & 0.09\\ 
    (2) & CF-3DGS + VFI & 31.45 & 0.93 & 0.09\\
    \midrule
    (3) & Baseline & 31.50 & 0.93 & 0.09\\
    (4) & + Global Training & 31.52 & 0.93 & 0.08\\
    (5) & + Progressive Training & 32.82 & 0.95 & 0.07 \\
    (6) & + Hierarchical Training (HT) & 33.02 & 0.95 & 0.07 \\
    (7) & + HT + VFI & 33.37 & 0.95 & 0.07 \\
    (8) & + HT + VFI + Base (Ours) & \cellcolor{bgcolor}33.53 & \cellcolor{bgcolor}0.96 & \cellcolor{bgcolor}0.07\\
    \midrule
    (9) & Ours w/o intrinsic & 32.17 & 0.94 & 0.09 \\
    \bottomrule
  \end{tabular}
  }
  \caption{\textbf{Ablation study on Tanks and Temples~\cite{Knapitsch2017}.} Incorporating all components yields the best performance. 
  }
  \label{tab:abs_overview}
  \vspace{-0.5cm}
\end{table}

\subsection{Ablation study}\label{sec:ablation_study}

We conduct an ablation study on the Tanks and Temples dataset~\cite{Knapitsch2017}. Table~\ref{tab:abs_overview} reports the average PSNR, SSIM, and LPIPS across all scenes, with scene-specific results available in the supplementary material. 
The baseline generally follows CF-3DGS~\cite{Fu_2024_CVPR}, with adjustments to certain hyperparameters to align with our strategy. \\

\noindent \textbf{Progressive vs. hierarchical training.} Table~\ref{tab:abs_overview} shows that hierarchical training (HT) outperforms progressive training (PT) by 0.2 dB (Variant 5 vs. 6), with both improving PSNR over the baseline (Variant 3) by more than 1.32 dB. This supports our claim that merging 3D Gaussians from different base models enhances results. 
HT performs better by balancing training across early and late frames, whereas PT, which allocates more iterations to early frames, tends to overfit them. However, PT is suitable for online tasks, unlike HT. 
Moreover, to test if the performance gains result from additional training iterations, we retrain the baseline model on the entire input (referred to as global training, GT). GT shows minimal improvement, indicating that retraining alone does not improve the learning of 3D Gaussians. \\

\noindent \textbf{Hierarchical training level.} 
We evaluate the effectiveness of hierarchical training at various levels $L$, with results in Table~\ref{tab:abs_hierarchical_level}. At $L\!=\!0$, the 3DGS model is trained by treating the entire input as a single segment. Our strategy notably boosts PSNR by 1.19–1.58 dB and SSIM by 0.02, while reducing LPIPS by 0.01–0.02. Performance increases with higher levels, saturating at $L\!=\!2$, where the video is divided into eight segments, sufficient for 3D Gaussian learning on the about 150-frame Tanks and Temples dataset.
Our proposal also lowers memory storage by 0.22–0.26 GB.
While it may seem counterintuitive, hierarchical training reduces memory storage for two reasons: (1) pruning unimportant Gaussians before merging, maintaining a stable count; (2) optimized segment-wise Gaussians are more representative, unlike the baseline model, which redundantly clones and splits Gaussians in regions with sparse Gaussians. \\

\noindent \textbf{Effectiveness of video frame interpolation (VFI).} 
We evaluate the impact of VFI in Table~\ref{tab:abs_overview}. VFI smooths camera motion and provides additional supervision, resulting in an average PSNR gain of 0.17dB (Variant 1 vs. 2) and 0.35 dB (Variant 6 vs. 7). VFI enhances performance across different 3DGS models, including CF-3DGS, not just within our hierarchical training pipeline. Moreover, hierarchical training contributes a PSNR gain exceeding 1.52 dB, showing that the core improvement stems from hierarchical training rather than augmented data from VFI. \\

\noindent \textbf{Effectiveness of supervision from base 3DGS models.}
Table~\ref{tab:abs_overview} shows that the supervision from base 3DGS models enhances performance, with an average PSNR increase of 0.16 dB and an SSIM increase of 0.01 (Variant 7 vs. 8). 
Pseudo-views generated by base 3DGS models mitigate overfitting and provide additional supervision for views not covered by the training images. \\

\noindent \textbf{Unknown camera intrinsics.} We experiment with heuristics instead of known camera intrinsics by setting the FoV to $70^{\circ}$. As shown in Table~\ref{tab:abs_overview}, PSNR dropped from 33.53dB to 32.17dB (Variant 8 vs. 9). Inaccurate camera intrinsics hinder pose estimation and may introduce scale ambiguity. Nonetheless, as seen in Table~\ref{tab:abs_overview}, our method, even without known intrinsics, outperforms CF-3DGS by 0.89dB (Variant 1 vs. 9).

\section{Conclusion}
We propose a hierarchical training strategy for 3D Gaussian splatting without known camera poses or SfM preprocessing, merging segment-specific base 3DGS models for enhanced representation.
We further incorporate video frame interpolation to smooth camera motion and mitigate overfitting by reusing interpolated images and base models. This approach outperforms state-of-the-art SfM-free novel view synthesis methods, enabling broader generalization across datasets without SfM preprocessing.  \\ 

\noindent \textbf{Limitations.} 
Our approach requires 
longer training
and can face challenges with large camera motion or low-quality inputs. While training time increases, rendering is faster due to fewer 3D Gaussians. In practice, training time can be reduced by lowering iterations or removing VFI, which is less necessary with small camera motion or abundant input frames. Large motion or poor inputs may cause alignment errors in 3DGS model merging.

{
    \small
    \bibliographystyle{ieeenat_fullname}
    \bibliography{main}
}


\end{document}